\documentclass[letterpaper, 10 pt, journal, twoside]{ieeetran}                   

\usepackage[bookmarks=true]{hyperref}
\usepackage{cite}
\usepackage{amsmath,amssymb,amsfonts}
\usepackage{graphicx}
\usepackage{gensymb}
\usepackage{textcomp}
\usepackage{xcolor}
\usepackage{placeins}
\usepackage{titlesec}
\usepackage{comment}
\usepackage{wrapfig}
\usepackage{placeins}
\usepackage[caption=false, font=footnotesize]{subfig}
\usepackage{multirow}
\usepackage{mathtools}
\usepackage[nolist,nohyperlinks]{acronym}
\usepackage{colortbl}
\usepackage{pdfpages}
\usepackage{algorithm}
\usepackage{algpseudocode} 
\usepackage[skip=2pt,font=footnotesize]{caption}
\usepackage{enumitem}
\setlist{nosep}

\usepackage{amsmath}

\usepackage{enumitem,amssymb}

\newlist{todolist}{itemize}{2}
\setlist[todolist]{label=$\square$}
\usepackage{pifont}

\newcommand{\revision}[1]{\textcolor{blue}{#1}}

\usepackage[capitalise]{cleveref} 

\crefname{algorithm}{algorithm}{algorithms} 
\Crefname{algorithm}{Algorithm}{Algorithms} 

\begin{document}

\acrodef{qp}[QP]{Quadratic Programming}

\title{Efficient Knowledge Transfer for Jump-Starting Control Policy Learning of Multirotors through Physics-Aware Neural Architectures}

\author{Welf Rehberg, Mihir Kulkarni, Philipp Weiss and Kostas Alexis%
\thanks{Manuscript received: October, 01, 2025; Revised December, 09, 2025; Accepted February, 13, 2026.}%
\thanks{This paper was recommended for publication by Editor Giuseppe Loianno upon evaluation of the Associate Editor and Reviewers' comments.
This work was supported by the Horizon Europe Grant Agreement No. 101119774 and the NVIDIA Academic Grant Program using NVIDIA RTX PRO 6000 Blackwell Max-Q and A100 GPU-Hours on Saturn Cloud. \textit{(Corresponding author: Welf Rehberg)}}%
\thanks{All authors are with the Department of Engineering Cybernetics at the Norwegian University of Science and Technology, O.S. Bragstads Plass 2D, 7034, Trondheim, Norway ({e-mails: \{\tt\footnotesize welf.rehberg, mihir.kulkarni, philipp.weiss, konstantinos.alexis\}@ntnu.no}).}%
\thanks{Digital Object Identifier (DOI): see top of the page.}
}

\markboth{IEEE Robotics and Automation Letters. Preprint Version. Accepted February, 2026}
{Rehberg \MakeLowercase{\textit{et al.}}: Efficient Knowledge Transfer for Jump-Starting Control Policy Learning of Multirotors}

\maketitle

\begin{abstract}

Efficiently training control policies for robots is a major challenge that can greatly benefit from utilizing knowledge gained from training similar systems through cross-embodiment knowledge transfer. In this work, we focus on accelerating policy training using a library-based initialization scheme that enables effective knowledge transfer across multirotor configurations. By leveraging a physics-aware neural control architecture that combines a reinforcement learning-based controller and a supervised control allocation network, we enable the reuse of previously trained policies. To this end, we utilize a policy evaluation-based similarity measure that identifies suitable policies for initialization from a library. We demonstrate that this measure correlates with the reduction in environment interactions needed to reach target performance and is therefore suited for initialization. Extensive simulation and real-world experiments confirm that our control architecture achieves state-of-the-art control performance, and that our initialization scheme saves on average up to $73.5\%$ of environment interactions (compared to training a policy from scratch) across diverse quadrotor and hexarotor designs, paving the way for efficient cross-embodiment transfer in reinforcement learning.

\end{abstract}


\section{Introduction}

\IEEEPARstart{M}{ultirotors} are popular for their agility and simple design, with recent advances in onboard autonomy leading to deployments in diverse environments~\cite{pretto2020building,chung2023into,dharmadhikari2025semantics} and even outperforming expert-level pilots in drone racing~\cite{kaufmann_champion-level_2023}. Yet, design still follows the traditional sequence: airframe first, then control policy. An emerging paradigm frames autonomous multirotor design as a computational co-design problem, jointly optimizing airframe and control policy \cite{gupta_embodied_2021, mannam_design_2024,park2021computational}. This approach envisions evaluating thousands of candidate designs by training a policy for each, but is fundamentally limited by the prohibitive cost of training so many policies. Training reinforcement learning (RL) policies is computationally intensive, and thus, suitable initialization can greatly accelerate convergence. A promising strategy is to leverage policies pre-trained on related configurations, either to initialize parameters \cite{julian_never_nodate} or guide exploration \cite{uchendu_jump-start_2023}. This relies on control policies generalizing across similar systems, a premise supported, among others, by domain randomization, which produces controllers effective not only for a nominal design in simulation but also for nearby system variations in reality.

\begin{figure}
    \vspace{-8pt}
    \centering
    \includegraphics[width=0.9\linewidth,clip]{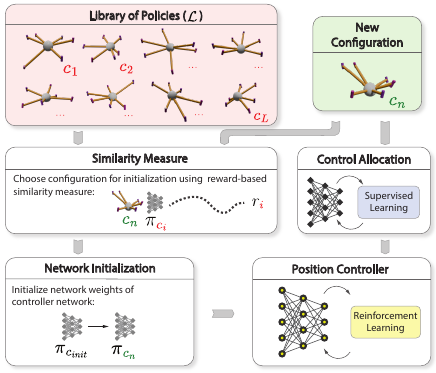}
    \caption{The proposed library-based initialization scheme: For training a policy for a new configuration, we first train a configuration-specific control allocation network. Subsequently, a suitable policy for initializating the training of the control policy is picked from a library of policies using a reward-based similarity measure. The resulting training time is significantly lower than training a policy from scratch.}
    \label{fig:lib_approach}
    \vspace{-8pt}
\end{figure}

We propose an efficient knowledge-transfer procedure for policy learning across diverse multirotor configurations, enabling fast training of large numbers of robots within the same system family. This challenge is not unique to co-optimization and also arises in general control-policy learning settings \cite{eschmann2025raptorfoundationpolicyquadrotor}. Our physics-aware neural architecture separates control into two parts: an RL-trained feedback policy mapping states to wrench vectors, and an allocation network trained via supervised learning guided by a \ac{qp} expert. This physics-aware architecture enables separate verification of controller and control allocation and supports reusing policies to jump-start learning for new airframe designs. To accelerate learning, we maintain a library of airframe–policy pairs. Using a similarity measure, a suitable prior configuration is selected whose policy and optimizer states are used to initialize training for new airframes. In turn, this greatly reduces the RL effort of training the controller, while the allocation network is always exclusively trained for the system at hand. We validate our approach through simulations, ablation studies, and real-world experiments, verifying its core contributions:
\begin{itemize}
    \item A library-based initialization scheme for control policy learning\revision{,} saving on average up to $73.5\%$ of environment interactions (compared to training a policy from scratch) across diverse quadrotor and hexarotor designs
    \item A physics-aware neural control architecture, facilitating policy initialization and improved interpretability 
\end{itemize}
The remainder of this paper is structured as follows: \cref{sec:related_work} reviews related work; \cref{sec:approach} presents the methods, including the multirotor dynamics, policy architecture, and initialization scheme; \cref{sec:experiments} reports real and simulated evaluations; and \cref{sec:conclusion} provides a summary and outlook.

\FloatBarrier

\section{Related Work}
\label{sec:related_work}
\vspace{-0.1cm}
Work on cross-embodiment policy transfer has focused mainly on manipulation tasks. Chen et al. \cite{chen_mirage_2024} propose zero-shot transfer, while Wang et al. \cite{wang2024crossembodimentrobotmanipulationskill} retrain network components to align action and state spaces. Uchendu et al. \cite{uchendu_jump-start_2023} reuse suboptimal policies as guides in a curriculum to accelerate exploration, and Julian et al. \cite{julian_never_nodate} show that fine-tuning policies initialized from other tasks reduces training time. While these works demonstrate transfer across configurations, they do not address how to select an initialization policy.
Earlier work on embodiment transfer in design optimization includes training general policies for efficient initialization \cite{luck_data-efcient_nodate,chen_pretraining-finetuning_2024}, transferring via intermediate embodiments \cite{liu_revolver_2022,mannam_design_2024}, and selecting from a discrete set of pretrained policies \cite{mannam_design_2024,le_goff_morpho_2023}. General-policy initialization requires either pretraining a broad policy \cite{chen_pretraining-finetuning_2024} or training multiple networks in parallel \cite{luck_data-efcient_nodate}, with both approaches constrained by the need for identical network architectures, limiting individual policy size. While general policies for multirotor control have not, to our knowledge, been applied to design optimization, they exhibit similar shortcomings \cite{eschmann2025raptorfoundationpolicyquadrotor} and are targeted to planar quadrotors so far \cite{eschmann2025raptorfoundationpolicyquadrotor, Zhang_2025}.
While using a curriculum of transitioning configurations in training is able to transfer policies for complex systems and tasks like object manipulation with robotic hands \cite{liu_revolver_2022}, they do not mention how the policy for initialization is picked among multiple already trained configurations, and the method adds significant additional complexity to the training process by making continuous changes to the robot model necessary. 
Earlier works on library-based approaches \cite{le_goff_morpho_2023, mannam_design_2024} mitigate these issues by using the original policies trained for individual configurations and reusing the experience gained with already trained policies. While \cite{le_goff_morpho_2023} reuses the best policy for each actuator-sensor combination, they do not consider the dynamical differences due to the different placement and orientation of the actuators and sensors. In \cite{mannam_design_2024}, the authors take the differences into account by approximating how similar two configurations are based on the difference of the parameter vector fully defining the configuration. Although this method offers a computationally efficient way to approximate how suitable a configuration is for initialization, it does not necessarily correlate well with improvements in sample efficiency when used for policy selection, as we will demonstrate later in this work. Additionally, no evaluation of the initialization scheme was conducted.
Last, we note that prior work on reinforcement learning for multirotors \cite{eschmann_learning_2024, hwangbo_control_2017} often uses end-to-end policies mapping states directly to motor commands. Drawing inspiration from classical control, we propose a modular framework in which one module generates wrench commands while another allocates actuator inputs. This separation enhances interpretability, particularly for arbitrary multirotor configurations. Building on this architecture, we further introduce a library-based initialization scheme that leverages a similarity measure and present an evaluation of its efficiency. 

\vspace{-7pt}
\enlargethispage{-1.0cm}
\section{Approach}
\label{sec:approach}

\subsection{Dynamical Systems Representation}
\label{sec:dyn_systems}
The morphology of the individuals of the multirotor family considered in this work is defined by the parameter vector: 
\begin{gather}
    c= [[t_1,...,t_{n_m}],[vec(\mathbf{R}_1),...,vec(\mathbf{R}_{n_m})]].
\end{gather}
with $n_m \in \mathbb{R}$ being the motor number and $t_i \in \mathbb{R}^{3}$ and $\mathbf{R}_i \in SO(3)$ being translations and orientations of the $i$-th rotor with respect to the body frame $\mathcal{B}$ placed at the center of mass as shown in \cref{fig:arbitrary_airframes}. $vec()$ describes flattening the lower-triangle matrix into a vector. In this work, we consider systems with $n_m \in \{4,6\}$ (i.e., quadrotors and hexarotors). The system state is described by $x = [p, v,q,\omega_B]^T \in \mathbb{R}^{13}$, with $p \in \mathbb{R}^{3}$ being the position, $v \in \mathbb{R}^{3}$ being the velocity (both expressed in the world frame $\mathcal{W}$), $q \in \mathbb{H}$ being the unit quaternion rotation (parametrizing the rotation matrix $\mathbf{R}(q)$), and $\omega_B \in \mathbb{R}^{3}$ being the rotational velocity in the body frame. The system dynamics can therefore be written as \cite{penicka_learning_2022}:

\begin{figure}
    \vspace{0.cm}
    \centering
    \includegraphics[width=0.52\linewidth]{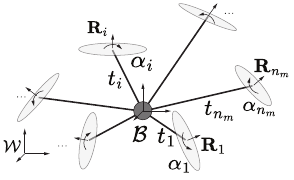}
    \caption{Description of arbitrary airframes considered in this work.}
    \label{fig:arbitrary_airframes}
    \vspace{-0pt}
\end{figure}

\begin{center}
  \begin{minipage}[b]{.2\textwidth}
  \begin{equation}
      \dot{p} = v 
      \label{eq:p_dot}
  \end{equation}
  \begin{equation}
      \dot{v} = \frac{1}{m}\mathbf{R}(q)F + g
      \label{eq:v_dot}
  \end{equation}
  \end{minipage}
  \quad
  \begin{minipage}[b]{.255\textwidth}
    \begin{equation}
      \dot{q} = \frac{1}{2}q \otimes 
      \left(\begin{array}{c} 
      0 \\
      \omega_B  \\
      \end{array}\right)
      \label{eq:q_dot}
    \end{equation}
    \begin{equation}
      \dot{\omega}_B = \mathbf{J}^{-1}(\tau - \omega_B \times \mathbf{J} \omega_B).
      \label{eq:om_dot}
    \end{equation}
  \end{minipage}
\end{center}
where $\mathbf{J} \in \mathbb{R}^{3 \times 3}$ denotes the inertia matrix of the system in $\mathcal{B}$ depending on the motor positions, $m \in \mathbb{R}$ its mass, $g = [0,0,-9.81~m/s^2]^T$ the gravity vector, $F \in \mathbb{R}^3$ the applied combined force, $\tau \in \mathbb{R}^3$ the applied torque in $\mathcal{B}$, and $\otimes$ denotes the quaternion product. We assume all systems to have the same payload and assume that the mass difference resulting from different arm lengths is negligible. The applied thrust and torque are related to the commanded motor thrusts $u \in \mathbb{R}^{n_m}$ as follows:
\begin{gather}
\left[ \begin{array}{c}F  \\
\tau  \end{array} \right]= \mathbf{F} u,
\end{gather}
\vspace{-0.5cm}
\begin{multline}
    \mathbf{F}=\left[
  \begin{matrix}
    \mathbf{R}_1  z_{m_1} & ...\\ 
    t_1 \times \mathbf{R}_1 z_{m_1} - \alpha_1 c_{q}\mathbf{R}_1 z_{m_1} &  ...  
  \end{matrix}\right.                
\\
  \left.
  \begin{matrix}
    \mathbf{R}_{n_m}  z_{m_{n_m}}\\ 
    t_{n_m} \times \mathbf{R}_{n_m} z_{m_{n_m}} - \alpha_{n_m} c_{q}\mathbf{R}_{n_m} z_{m_{n_m}}
  \end{matrix}\right].
\end{multline}
with $\mathbf{F} \in \mathbb{R}^{6\times n_m}$ representing the allocation matrix of the system, $\alpha_i\in{\{1,-1\}}$ representing the rotation direction of rotor $i$, $c_q \in \mathbb{R}$ being the torque to thrust ratio of the motors and $z_{m_i} \in \mathbb{R}^3$ representing the thrust direction of motor $i$ in motor frame ($[0,0,1]^T$ for all motors). The relation between the commanded motor thrust and the rotational velocity ($\omega_i$) of motor $i$ is defined by $u_i = c_t \omega_i^2$ with $c_t \in \mathbb{R}$ being the thrust coefficient.
\subsection{Physics-aware Control Policy Learning}
\label{sec:rl}
\begin{figure}
    \vspace{0pt}
    \centering
    \includegraphics[width=\linewidth]{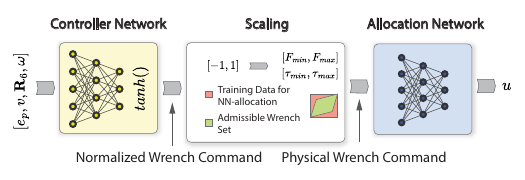}
    \caption{Proposed physics-aware neural control architecture. The architecture is split into a controller and an allocation network, which are trained separately. The control network is trained using RL, while the allocation network is trained via supervised learning.}
    \label{fig:neuralarchitecture}
    \vspace{-8pt}
\end{figure}

In this work, we depart from earlier efforts on learning position controllers that directly output motor commands through a single neural network~\cite{eschmann_learning_2024, hwangbo_control_2017} and propose a two-step approach that employs a physics-aware neural architecture trained with a combination of reinforcement and supervised learning. The proposed neural architecture is depicted in ~\cref{fig:neuralarchitecture}. As shown, the method involves first learning the control allocation (from desired wrench to motor commands) in a supervised manner and then fixing that learned network while training a position controller outputting wrenches ($w = [F,\tau]^T$) through reinforcement learning.
Splitting the controller into two parts leaves us with a network where the first part depends on the admissible wrench set, the mass, and the inertia of the system, while the second part is specific to the system's allocation matrix. Since the first part of the controller commands forces and torques to a rigid body, this is hypothesized and demonstrated in practice that it generalizes between different system configurations (including different motor numbers) and can, therefore, be initialized efficiently from already trained policies for other configurations. Additionally, the state between the two networks (wrench) is physically meaningful, thus adding interpretability to the controller.

\subsubsection{Learning-based Control Allocation}
\begin{figure}
    \vspace{0pt}
    \centering
    \includegraphics[width=.5\linewidth]{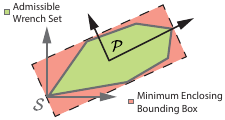}
    \caption{Transformation of wrench commands.}
    \label{fig:command_transform}
    \vspace{-8pt}
\end{figure}

Learning the control allocation, compared to using the pseudo-inverse of the allocation matrix or solving an optimization problem, comes with three advantages. First, compared to pseudo-inverse methods that find a solution solving the unconstrained allocation \cite{tognon_omnidirectional_2018}, the learned control allocation can be trained to mimic an expert that accounts for constraints as described in this section. This is increasingly important for systems with a low thrust-to-weight ratio. Second, solving the constrained allocation optimization problem at each timestep during training slows down the training process (solving the optimization problem is significantly more expensive than the forward path through a network), which is especially problematic when large numbers of different systems have to be trained. Third, at runtime, performing inference through that network comes with predictable computation time as opposed to algorithms that iteratively solve for the constrained control allocation~\cite{boyd_convex_2023}. To learn the control allocation, we generate training data by sampling wrenches uniformly from a hypercube containing the admissible set of the configuration. Choosing the space from which the training data is sampled to be a hypercube is motivated by the fact that normalized commands (between $-1$ and $1$) generated by the control network can be mapped directly (without solving a linear problem) onto the training envelope, guaranteeing the allocation network is not confronted with data unseen in training. Our experiments have shown that the achievable accuracy of the allocation network is highly dependent on the hyper-volume ratio between the chosen sample space and the admissible set. To minimize this ratio, we calculate an axis-aligned minimum enclosing bounding box that is aligned with the principal components of the vertices of the convex hull of the admissible set. During inference, the normalized commands of the controller network are scaled according to the limits of the bounding box in the frame $\mathcal{P}$ defined by the principal components as shown in \cref{fig:command_transform}. Then, the commands are transformed from $\mathcal{P}$ to the frame $\mathcal{S}$ of the admissible wrench space.

For generating the corresponding motor forces from the sampled wrenches, we formulate the control allocation as a slack-constrained quadratic program following \cite{johansen_control_2013}, minimizing projection error and motor forces:
\begin{gather}
\begin{array}{rrclcl}
\displaystyle \min_{u,s} & ||u|| + ||s||\\
\textrm{s.t.} & u_{min} \leq u_i \leq u_{max} \forall i \\
& \mathbf{F}u-w_d = s.
\end{array}
\end{gather}
with $w_d \in \mathbb{R}^6$ being the commanded wrench, $u_{min}\in \mathbb{R}^{n_m}$ and $u_{max}\in \mathbb{R}^{n_m}$ being the minimum and maximum motor thrust and $s \in \mathbb{R}^6$ being a slack variable. Introducing $s$ is necessary since it is not guaranteed that the commanded wrench is within the admissible wrench set. The optimization problem is solved for multiple data points in parallel using the \texttt{qpth-solver} \cite{amos_optnet_2021} to speed up the data generation. The allocation network is a simple MLP with different layer sizes, network depth and activation function depending on whether constraints should be enforced. Since a network imitating the unconstrained control allocation QP is equivalent to the solution obtained with the pseudo-inverse, small layer sizes are sufficient and no non-linear activation functions are needed. For unconstrained control allocation, a network with one layer and $32$ neurons is chosen. Introducing constraints renders the problem highly non-linear and requires larger networks and non-linear activations. Therefore, a network with $3$ layers with $100$ neurons each and $Leaky~ReLU$ as an activation function is chosen.   

\subsubsection{Learned Position Controller}
For training the controller network, we formulate the control problem as a Markov Decision Process (MDP) using reinforcement learning. As input, the network receives the state representation $x$ where we replace the position with the position error $p_e$ and the quaternions with the 6-D rotation representation ($R_{6D}$) introduced in \cite{zhou_continuity_2020} to avoid double coverage of $SO(3)$ without introducing redundancy. This results in the observations $o=[p_e,v,R_{6D},\omega] \in \mathbb{R}^{15}$.

The used reward is based on the following quantities:
\begin{align}
    h_{p} &= [p_{e_x},p_{e_y},p_{e_z}], ~h_{v} = [v_x,v_y,v_z]\label{eq:rew_p},\\
    h_{\Omega} &= [\omega_x,\omega_y,\omega_z]\label{eq:rew_ang_vel},\\
    h_{u_{abs}} &= [u_1 - \hat{u}_1,...,u_4 - \hat{u}_4]\label{eq:rew_u_abs},\\
    h_{up} &= 1 - (q e_3 q^{-1})_z, ~h_{forw} = 1 - (q e_1 q^{-1})_x.
    \label{eq:rew_forw}
\end{align}
The quantities $h_p$, $h_v$ and $h_{\Omega}$ are terms based on the position offset from the provided reference, the linear velocity and the angular rates. The term $h_{u_{abs}}$ is based on the difference between each commanded motor thrust and the thrust necessary for hovering ($\hat{u}$). To incentivize smooth actions, we add temporal smoothness regularization terms and spatial smoothness terms to the policy loss as described in \cite{mysore_regularizing_2021}. The quantity $h_{up}$ is based on the alignment of the $z$-axis of $\mathcal{B}$ and the $z$-axis of $\mathcal{W}$. This is calculated by rotating the corresponding unit vector $e_3$ by the quaternion representing the orientation of the system $q$ and evaluating its alignment, where $h_{up} = 0$ for perfect alignment. $h_{forw}$ is equivalently based on the alignment of the $x$-axis of $\mathcal{W}$ and $\mathcal{B}$. Note that strictly speaking, only $h_p$ is relevant for the position tracking task and the other terms are only to ensure a desired behavior from a sim-to-real standpoint. The terms in equation \ref{eq:rew_p} - \ref{eq:rew_forw} described quantities $h$ which are used to calculate the individual parts $r_k$ of the reward function using exponential kernels as follows:
\begin{align}
    r_k =  \sum_i ae^{-bh_i^2},
\end{align}
where the coefficients $a \in \mathbb{R}$ and $b \in \mathbb{R}$ of the exponential kernel are adequately chosen weights defining its magnitude and width and $k$ indicates the quantity the reward is based on ($p,v,\Omega,...$). The obtained individual reward terms are combined into the following cumulative reward: 
\begin{align}
    r = r_p(r_{forw} + r_v + r_{\Omega}) + r_p + r_v + r_{up} + r_{{a}_{abs}}.
\end{align}
Multiplying $r_{forw}$, $r_v$ and $r_{\Omega}$ with $r_p$ leads to a weighted reward allowing for high linear velocities, angular rates and orientation offsets far away from the target location. The controller network consists of two layers with $32$ and $24$ neurons with $tanh$ as activation function. Additionally, we employ $tanh$-squashing after the last layer to bound the generated outputs to a $[-1,1]$-hypercube. During training, no specific trajectories are sampled. Instead, we sample an initial state of the system and the reward function penalizes deviations from the origin of the state space.
\subsection{Initializing Policies to Jump-Start Policy Learning}
\label{sec:initializing_policies}
To accelerate the training of a new configuration, we propose a library-based initialization scheme shown in \cref{fig:lib_approach} and \Cref{alg:lib_based_init}, maintaining a fixed number of candidate policies. For a new configuration defined by the parameter vector $c_n$, first, a network for control allocation is trained. To pick the initial weights for the controller network, all configurations $c_k \in \mathcal{L}=\{c_1,...,c_L\}$ in the library $\mathcal{L}$ are evaluated regarding their suitability for initialization based on a similarity measure $m(c_i,c_k)$. The policy weights of the configuration with the best score are picked as initial values for the training of the new configuration. Initializing the actor weights alone is not sufficient and will result in a poor learning signal provided by the untrained critic, as described in \cite{uchendu_jump-start_2023}. We therefore initialize the critic in the same way as the actor, similar to \cite{chen_pretraining-finetuning_2024}. Additionally, we found that transferring the current optimizer state (first and second moments of the gradients for Adam) of the policy used for initialization resulted in significantly better performance. Subsequently, the configuration is fine-tuned until convergence. 
\vspace{-0.3cm}
\small
\begin{algorithm}
  \caption{Library-Based Initialization}
  \label{alg:lib_based_init}
  \begin{algorithmic}
    \State \textbf{Given:} Parameter vector of the configuration to train $c$, library of already trained configurations $\mathcal{L}$, similarity measure \texttt{m($c_i$,$c_k$)}, RL training method \texttt{train($c$,$\Theta$,$\gamma$)}.
    \vspace{5pt}
    \State initialize empty list $M$
    \For{$c_{k}$ in $\mathcal{L}$}
        \State Append $m(c,c_k)$ to $M$ \Comment{compute similarity measure }
    \EndFor
    \State $c_{init} = \mathcal{L}(\arg \max(M))$ \Comment{pick config for initialization} 
    \State \quad \quad \quad \quad \quad \quad \quad \quad \quad \quad \quad \quad ($\arg \min$ for $m_c,m_{wd}$)
    \State $\Theta = \Theta_{init}$ \Comment{initialize policy parameters}
    \State $\gamma = \gamma_{init}$ \Comment{initialize optimizer states}
    \State \texttt{train($c$,$\Theta$,$\gamma$)} \Comment{train config $c$ until convergence}
  \end{algorithmic}
\end{algorithm}
\normalsize
\vspace{-0.1cm}
Our approach requires an efficient method for selecting a policy for initialization from the library. In this work, two avenues are explored to select a configuration for initialization: first, selecting a configuration based on a measure over the physical properties of the airframe, and second, evaluating the properties of the policy. This led to the $3$ following similarity measures considered in this work:

\begin{enumerate}
     \item A straightforward choice for comparing the physical properties of two systems is to evaluate the norm of the difference of their augmented parameter vectors, like in \cite{mannam_design_2024}. The augmented parameter vector is defined as $c_{aug} = [c,vec(\mathbf{J})]$ and the corresponding measure is defined as:
     \vspace{-0.4cm}
     \begin{gather}
         m_c(c_{aug_i},c_{aug_k}) = ||c_{aug_i}-c_{aug_k}||_2.
     \end{gather}
     \vspace{-0.4cm}

     \item The admissible set of linear and angular accelerations that a configuration can produce instantaneously closely relates to the interface between the policy and the system, while implicitly providing information on all physical properties of the airframe (mass, inertia, and motor position and orientation). However, comparing the admissible acceleration sets for two arbitrary configurations is not straightforward (e.g., collapsing overlapping volume of the convex hulls). Instead, we propose a measure based on the Wasserstein distance \cite{panaretos_statistical_2019} between two distributions ($X, Y$) representing the admissible acceleration sets ($A(c)$):
     \begin{gather}
     X \sim A(c_i), \quad Y \sim A(c_k), \\
     m_{wd}(c_i, c_k) = \inf_{\pi}(1/n\sum^n_{j=1}||X_j-Y_{\pi(j)}||_2)^{(1/2)}.
     \end{gather}
     where the infimum is over all permutations $\pi$ with $n$ being the number of samples drawn from a uniform distribution bounded by the convex hull of the admissible set. Intuitively, the measure can be interpreted as the ``effort'' necessary to transform one set into the other.
     
     \item Instead of choosing a configuration based on the physical properties, we also explore directly evaluating the policies of the configurations in the library. To approximate how well a configuration is suited for initialization, we deploy its policy on the newly sampled configuration and evaluate the accumulated reward. Using modern massively parallelized simulators, this can be done for large numbers of configurations in parallel. The resulting measure can be formulated as:
     \vspace{-0.2cm}
     \begin{multline}
         m_r(c_i,c_k) = \mathbb{E}[\sum_{t=0}^Tr(x_{t},u_{t},c_i)|
         \\
         x_0=x_{init},u_t \sim \pi_{c_k}(x_t)].
     \end{multline}
     with $T$ being the number of evaluation time steps, $\pi_{c_k}$ being the policy corresponding to the configuration defined by the parameter vector $c_k$, and $x_{init}$ being the initial state.
     
\end{enumerate}

\section{Evaluation}
\label{sec:experiments}

To show the validity of our proposed initialization scheme and controller structure, we run multiple simulations and real-world experiments. First, we perform extensive simulation studies to evaluate the efficiency of our library-based initialization scheme and how it leads to computational improvements. Subsequently, we evaluate the accuracy of the learned control allocation networks by comparing them to the QP-expert used for training. Finally, we demonstrate the proposed controller structure, consisting of the controller network and the allocation network, in real-world scenarios.

\subsection{Evaluation of the Library-based Initialization Scheme}

The experiments to evaluate the proposed initialization scheme are twofold. First, the correlation of the similarity measures with the number of saved environment interactions are evaluated to decide which one should be used in the initialization scheme. Second, the initialization scheme based on the selected similarity measure is evaluated.

\begin{figure}
    \vspace{0pt}
    \centering
    \includegraphics[width=0.9\linewidth]{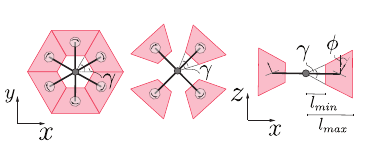}
    \caption{Sampling range for sampling the configurations in the pool. With $l_{min}=0.1m$, $l_{max} = 0.35m$, $\gamma = 60\degree$ and $\phi=20\degree$.}
    \label{fig:sampling_range}
    \vspace{-8pt}
\end{figure}

\begin{figure}
\vspace{-6pt}
    \centering
    \includegraphics[width=0.8\linewidth]{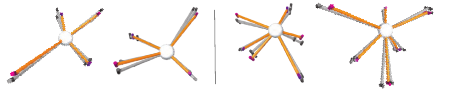}
    \caption{Example configurations from the sampled library of configurations with two of their closely sampled configurations in greyscale (quadcopters left and hexacopters right).}
    \label{fig:lib_examples}
    \vspace{-8pt}
\end{figure}

To evaluate the proposed initialization scheme, a library of $40$ configurations is sampled with $4$ motors and with $6$ motors, while a control policy providing wrench commands to an unconstrained control allocation network is trained for each configuration from scratch until convergence for $3$ different seeds. We choose to evaluate the initialization scheme using unconstrained networks because training the constrained control allocation takes significantly more time. The motor positions are uniformly sampled from a cone stump centered at the nominal motor position of the standard quad- and hexacopter. The two faces are defined by the maximum and minimum arm length ($l_{min}$, $l_{max}$), while the cone angle $\gamma$ defines the angle limits of the arm. The motor orientations are uniformly sampled such that the motor $z$-axis does not deviate from the body $z$-axis by more than $\phi$. This is illustrated by \cref{fig:sampling_range}. The high-dimensional sampling space induced by the large number of morphology parameters implies that dense sampling across all of it is computationally expensive. For the purposes of the presented evaluation, we restrict the sampling space to a localized region to obtain meaningfully close configurations for initialization. In order to still be able to evaluate diverse airframes, we sample $3$ additional configurations uniformly from a distribution centered around each of the $40$ configurations, allowing for a maximum deviation of $5\degree$ in motor angle and $5cm$ in motor position (examples shown in \cref{fig:lib_examples}). Those configurations are added to the library ($160$ configurations in total). Compared to simulating the real system, the configurations in the pool are simulated with a lower fidelity simulation. This includes calculating the system's inertia from a point mass model and removing sensor noise and the motor model from the simulation. Additionally, the reward function is condensed to include only terms that matter for the task of position control (purely based on the position error). The lower fidelity simulation and only using position terms facilitates using the same reward function across the whole system family. Additionally, we check that the sampled configurations are able to hover and have control authority around all axes. Configurations that are still not able to learn in any of the tried seeds are discarded after training.  

\subsubsection{Similarity Measures}

We assess the effectiveness of the proposed similarity measures in predicting whether a configuration is suitable for initialization by examining the correlation between the similarity measure and the change in the number of environment interactions necessary to reach a goal reward. To this end, we retrain every configuration in the pool while initializing the training with randomly picked configurations from the other configurations in the pool. We then compute the Spearman correlation coefficient and corresponding $p$-value to assess the relationship between the similarity measure calculated for two configurations and the resulting change in required environment interactions when one is used to initialize the other. Since with a flattening reward curve, the randomness in the training process has an increasing influence on the exact number of saved environment interactions, the correlation becomes less strong. We therefore evaluate the correlation and statistical significance at multiple reward goals over the full reward range and report the values for the highest reward a $p$-value $\leq0.01$ was obtained to ensure high significance. The corresponding results are summarized in \cref{tab:eval_sim_measure}. The results show that the two measures based on evaluating physical properties of the configuration correlate less well than the measure $m_r$ based on evaluating the policy directly. Note that the $p$-values of the evaluations of $m_c$ and $m_{wd}$ show no statistical significance and have a smaller correlation coefficient over the whole range.

\renewcommand{\arraystretch}{1.2} 
\begin{table}[t]
    \caption{Statistical evaluation of the similarity metric.}
    \centering
    \scriptsize
    \begin{tabular}{|c|c|c|c|c|}
        \hline
         \multirow{2}{*}{\begin{tabular}{@{}c@{}} \textbf{Similarity} \\ \textbf{Measure} \end{tabular}} &  \multicolumn{2}{c|}{\textbf{Abs. Corr. Coefficient}}  & \multicolumn{2}{c|}{$p$\textbf{-value}} \\
        \cline{2-5}
        & \quad \textit{Quad} \quad \quad & \textit{Hex} & \textit{Quad} & \textit{Hex} \\
        \hline
        \hline
          $m_c$ & $0.11$  & $0.23$ & $0.54$ &  $0.26$  \\
        \hline
          $m_{wd}$ & $0.39$ & $0.10$ & $0.03$ & $0.65$  \\
        \hline
        $m_r$ & $\mathbf{0.52}$  & $\mathbf{0.55}$ & $\mathbf{2.5\times10^{-3}}$ & $\mathbf{4.4 \times 10^{-3}}$ \\
        \hline
    \end{tabular}
    \normalsize
    \label{tab:eval_sim_measure}
\end{table}
\renewcommand{\arraystretch}{1.} 

\subsubsection{Initialization Scheme}
\begin{figure}
    \vspace{0pt}
    \centering
    \includegraphics[width=.8\linewidth]{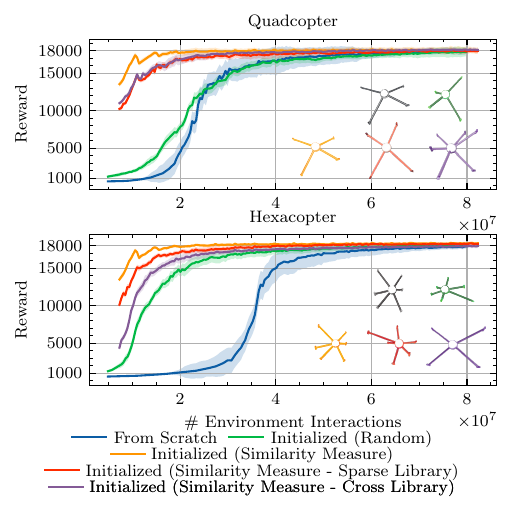}
    \caption{Results of the policy initialization experiments. The blue line shows training from scratch, the green line shows initialization with a random configuration, the red line shows initialization using the proposed similarity measure, selecting policies from the original $40$ configurations without the added close neighbors and the orange line shows initialization using the similarity measure, selecting policies from the full library of systems, including close neighbors. The violet line shows initialization using the proposed measure, picking policies for initialization from the library of systems with another motor number.}
    \label{fig:exp_initialization}
    \vspace{-8pt}
\end{figure}

 Subsequently, each configuration in the pool is retrained while picking a policy for initialization based on the reward-based similarity measure. All training runs were repeated using $3$ different seeds. To show that choosing a configuration based on the proposed similarity measure is important, we compare our library-based initialization scheme to randomly choosing configurations from the library. This shows that the constraints on the configuration space we impose do not render the configurations so close to each other that picking arbitrary configurations will yield similar results as using the proposed metric. Additionally, we conduct experiments initializing only from the original (sparse) library, not containing close neighboring configurations and initializing from policies trained for a different motor number. In \cref{fig:exp_initialization}, the median of the reward curves for all systems and the mean of the standard deviation over seeds for each configuration are reported. The experiments show that the proposed initialization scheme reduces required environment interactions by $\approx71/76\%$ (quadrotor/hexarotor) with close configurations, $\approx50/65\%$ with the sparse library, and $\approx49/65\%$ when using a policy from a system with a different motor number. Note that computing the similarity measure increases the environment interactions (accounted for in the figure), but choosing a system afterward is trivial. \cref{fig:exp_initialization} shows a significant difference when selecting a random configuration for initializing the hexacopter training. This difference can be attributed to the greater number of randomly sampled motors, which, on average, results in an admissible set that more uniformly covers the wrench space. As a result, the overlap between the admissible sets of the two configurations increases, facilitating smoother initialization.

\subsection{Evaluation of the Learned Control Allocation}
\begin{figure}
    \centering
    \includegraphics[width=0.98\linewidth]{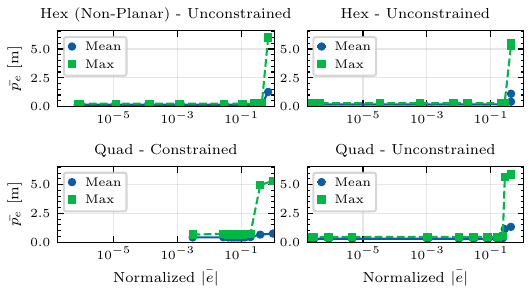}
    \caption{Closed-loop trajectory tracking error with respect to the mean absolute allocation error normalized to the thrust range of the networks.}
    \label{fig:sensitivity}
\end{figure}

In the following, the learned control allocation is compared to the QP-expert solving the constrained and unconstrained control allocation. For evaluation, $3$ configurations having $4$ motors and $3$ configurations having $6$ motors were used. For both motor numbers, the standard configuration (all thrust axes parallel), a configuration where all motors are tilted by $10^\circ$ outwards, and a random configuration were chosen. Table \ref{tab:proj_err_control_all} shows the average and maximum errors normalized to the thrust range between the output of the learned allocation model and the solution to the quadratic problem over $1e5$ random samples. In practice, this gives an intuition of how close the outputs of the controller network will be to physically meaningful wrenches. Since the unconstrained control allocation is a linear mapping, the network is able to reproduce the expert outputs with very high precision. The constrained control allocation, on the other hand, is a nonlinear relation requiring more complex networks. Our experiments show that the accuracy of the learned control allocation drops for higher motor numbers. This is likely due to the higher-dimensional output space (reflecting the actual control allocation of the system) and, therefore, more complex mapping. Learning the constrained control allocation can still be beneficial for arbitrary designs since it allows for constraint-aware allocation and, therefore, redistribution of effort in the case of saturation. The observed errors, particularly during training of the constrained control allocation network, do not hinder the controller network, as it learns to compensate for these inaccuracies by adapting its control strategy accordingly. Additionally, sensitivity curves of the closed-loop tracking performance of a frozen control network with respect to the allocation error are shown in \cref{fig:sensitivity}. It is observed that even without allowing the network to adapt to the allocation error, there is a significant margin before the allocation error significantly influences the closed-loop tracking error.

\begin{table}[]
    \centering
    \renewcommand{\arraystretch}{1.2} 
    \scriptsize
    \caption{Maximum ($|e|_{\max}$) and average ($\bar{|e|}$) absolute errors of the control allocation normalized to the thrust range}
    \begin{tabular}{|c|c|c|}
        \hline
          & \begin{tabular}{@{}c@{}}\textbf{Constrained} \\ $|e|_{\max}$, $\bar{|e|}$ \end{tabular} 
          & \begin{tabular}{@{}c@{}}\textbf{Unconstrained} \\ $|e|_{\max}$, $\bar{|e|}$ \end{tabular} \\
        \hline\hline

        Quadrotor & $2.7\mathrm{e}\!-\!2$, $1.2\mathrm{e}\!-\!3$ & $4.4\mathrm{e}\!-\!5$, $5.2\mathrm{e}\!-\!6$ \\
        \hline

        \begin{tabular}{@{}c@{}}Quadrotor with \\ Tilted Propellers ($10^\circ$)\end{tabular}
        & $2.5\mathrm{e}\!-\!2$, $1.3\mathrm{e}\!-\!3$ & $4.0\mathrm{e}\!-\!5$, $5.0\mathrm{e}\!-\!6$ \\
        \hline

        Random Quadrotor & $4.3\mathrm{e}\!-\!2$, $1.8\mathrm{e}\!-\!3$ & $8.4\mathrm{e}\!-\!5$, $8.1\mathrm{e}\!-\!6$ \\
        \hline

        Hexarotor & $9.0\mathrm{e}\!-\!2$, $6.0\mathrm{e}\!-\!3$ & $5.9\mathrm{e}\!-\!4$, $1.4\mathrm{e}\!-\!5$ \\
        \hline

        \begin{tabular}{@{}c@{}}Hexarotor with \\ Tilted Propellers ($10^\circ$)\end{tabular}
        & $15.0\mathrm{e}\!-\!1$, $7\mathrm{e}\!-\!3$ & $1.7\mathrm{e}\!-\!3$, $3.8\mathrm{e}\!-\!4$ \\
        \hline

        Random Hexarotor & $11.0\mathrm{e}\!-\!1$, $4\mathrm{e}\!-\!3$ & $8.7e\!-\!5$, $7.4\mathrm{e}\!-\!6$\\ 
        \hline
        
    \end{tabular}
    \label{tab:proj_err_control_all}
    \normalsize
    \vspace{-10pt}
    \renewcommand{\arraystretch}{1.}
\end{table}

\subsection{Evaluation of the Proposed Controller Architecture}

\begin{figure}
    \vspace{-0cm}
    \centering
    \includegraphics[width=0.85\linewidth]{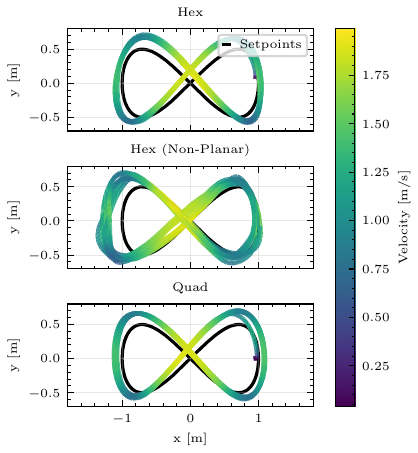}
    \caption{Results of the real-world trajectory tracking experiments using an unconstrained control allocation network.}
    \label{fig:network_obs_unconstrained}
    \vspace{-8pt}
\end{figure}

\begin{figure*}[t]
    \vspace{-2pt}
    \centering
    \begin{minipage}{0.55\linewidth}
        \includegraphics[width=\linewidth]{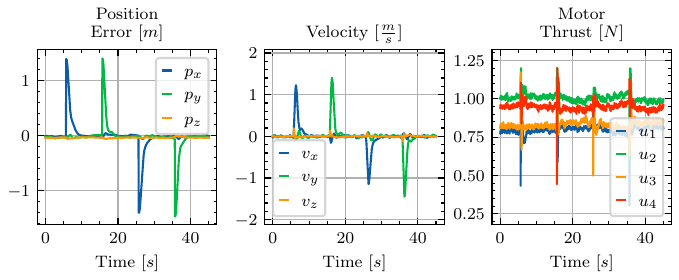}
    \end{minipage}
    \hspace{.5cm}
    \begin{minipage}{0.4\linewidth}
        \centering
        \includegraphics[width=\linewidth]{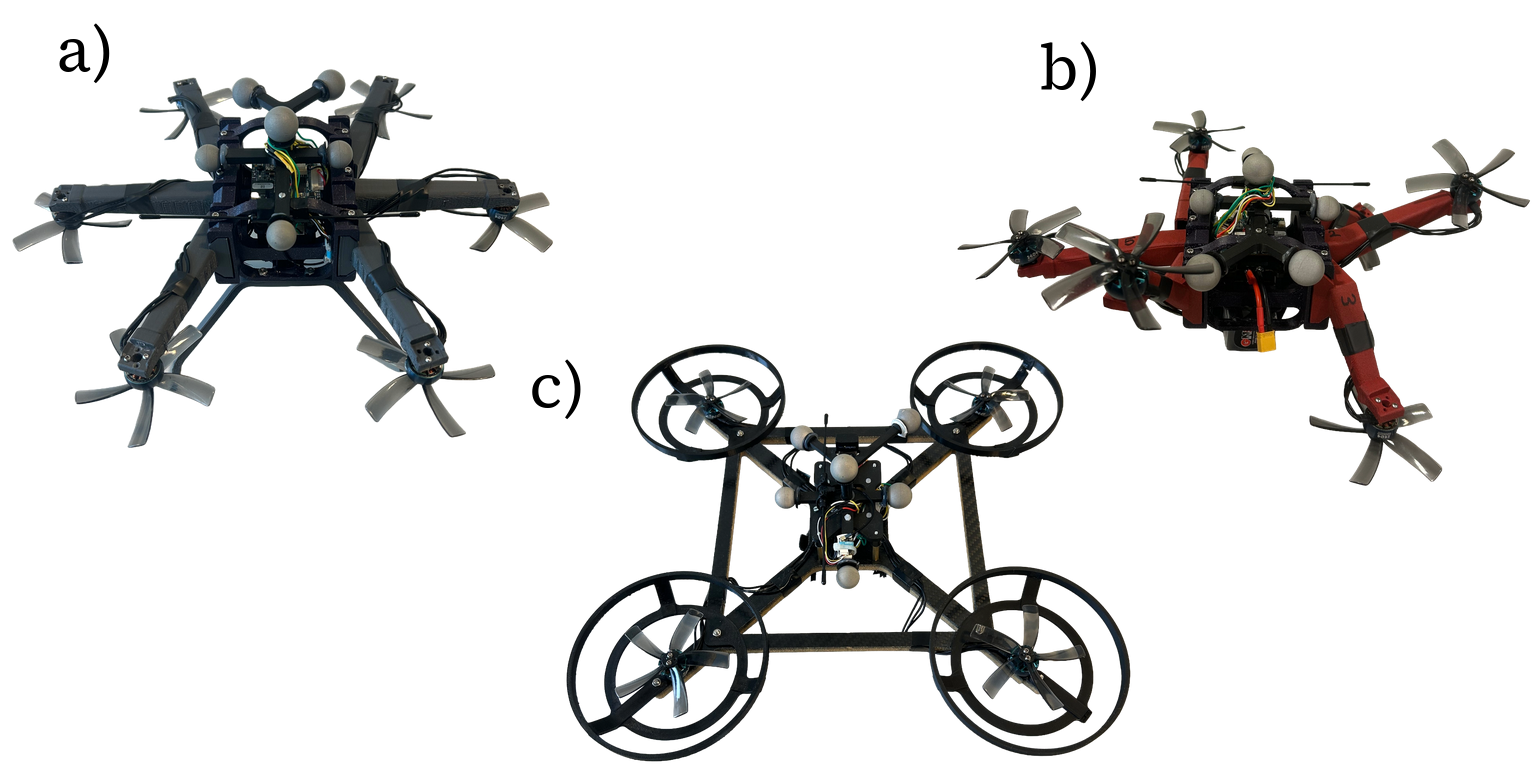}
    \end{minipage}
    \caption{Left: results of the real-world experiments using a constrained control allocation network. Right: systems used for real-world experiments ( a) planar hexarotor, b) non-planar hexarotor, c) planar quadrotor).}
    \label{fig:combined}
    \vspace{-8pt}
\end{figure*}

We train our RL policies with Proximal Policy Optimization (PPO) \cite{schulman_proximal_2017} using a customized version of Sample-Factory \cite{petrenko_sample_2020} and simulate the system in the Aerial Gym Simulator \cite{kulkarni_aerial_2025}. All policies are trained on a Lenovo ThinkPad P1 Gen 6 from 2024 with a GeForce RTX 4090 GPU and deployed on a system with a ModalAI Voxl 2 Mini board and a ModalAI Voxl ESC. To guarantee robust sim-to-real transfer, we incorporate sensor noise and highly accurate approximations of the robot's inertia obtained from the CAD model and the true robot mass into the simulator. We additionally simulate the motor dynamics using a first-order model with a time constant obtained from RPM step-response experiments on the real system. All policies are trained with a physics time step size of $0.01s$ and deployed with a control frequency of $250\textrm{Hz}$. The trained networks were deployed on the compute board with a custom PX4 module. We evaluate the proposed controller architecture in $2$ different experiments. First, we deploy a trained policy to track a Lissajous trajectory with a loop time of $5.5s$ for a planar hexacopter ($14cm$ arm length and a total weight of $421g$) and quadcopter ($23cm$ arm length and a total weight of $373g$) configuration and a non-planar system from the library ($14cm$ arm length on average and a total weight of $402g$) with an unconstrained allocation network. Note that we slightly adjust the non-planar system's arm angles and lengths to avoid self-collision and fit the arms to the body of an existing frame. For dynamic trajectories, the network receives the velocity error to a setpoint instead of the system velocity. The resulting trajectories are shown in \cref{fig:network_obs_unconstrained}. Subsequently, we deploy a trained policy for a standard quadcopter configuration with a constrained allocation network in a static setpoint tracking task (shown in \cref{fig:combined}). \cref{fig:combined} shows a constant offset in the commanded thrust values, which is not seen in simulation. This is likely due to unmodeled shifts in the center of mass or differing thrust coefficients between motors. We achieve state-of-the-art \cite{kulkarni_aerial_2025} steady state position error $\leq0.055m$ and a mean trajectory tracking error of $0.26m$ for the quadrotor, $0.17m$ for the planar hexacopter and $0.24m$ for the non-planar hexacopter.
 
\section{Conclusion and Future Work}
\label{sec:conclusion}
In this work, we presented a library-based policy initialization scheme aimed at accelerating the training of control policies for diverse multirotor configurations. By leveraging a physics-aware, modular control architecture and introducing a policy evaluation-based similarity measure, we enable the efficient cross-embodiment transfer of previously trained policies to jump-start learning for novel designs. Our results in simulation and the real world a) demonstrate substantial improvements in sample efficiency and training time and b) showcase that the proposed control architecture is capable of state-of-the-art performance. We further show that the similarity measure based on policy behavior significantly outperforms those based purely on physical configuration parameters in predicting initialization success. This highlights the importance of task-specific evaluation criteria for policy transfer. In the future, the presented work could be extended by initializing navigation policies for systems with differently placed exteroceptive sensors. 

\bibliographystyle{IEEEtran}
\bibliography{bibliography/Library}

\end{document}